\documentclass[acmtog]{acmart}

\acmSubmissionID{1502}
\usepackage{makecell}
\usepackage{bm}

\copyrightyear{2026}
\acmYear{2026}
\setcopyright{cc}
\setcctype{by}
\acmConference[SIGGRAPH Conference Papers '26]{Special Interest Group on Computer Graphics and Interactive Techniques Conference Conference Papers}{July 19--23, 2026}{Los Angeles, CA, USA}
\acmBooktitle{Special Interest Group on Computer Graphics and Interactive Techniques Conference Conference Papers (SIGGRAPH Conference Papers '26), July 19--23, 2026, Los Angeles, CA, USA}
\acmDOI{10.1145/3799902.3811218}
\acmISBN{979-8-4007-2554-8/2026/07}

\begin{document}

\title[Policy-based Foveated Imaging and Perception]{Policy-based Foveated Imaging and Perception}

\author{Howard Xiao}
\orcid{0009-0009-5256-8988}
\affiliation{%
  \institution{Stanford University}
  \country{USA}
}

\author{Jan Ackermann}
\orcid{0009-0006-7875-3135}
\affiliation{%
  \institution{Stanford University}
  \country{USA}
}

\author{Boyang Deng}
\orcid{0009-0008-5133-8008}
\affiliation{%
  \institution{Stanford University}
  \country{USA}
}

\author{Gordon Wetzstein}
\orcid{0000-0002-9243-6885}
\affiliation{%
  \institution{Stanford University}
  \country{USA}
}

\begin{abstract}
Ultra-high-resolution image sensors offer the potential to capture fine spatial details critical for many visual perception tasks, but acquiring and processing all pixels at full resolution is often infeasible under realistic bandwidth, latency, and power constraints. Existing approaches address this challenge through acquisition strategies such as spatial or temporal downsampling, which irrevocably discard information before task relevance can be assessed. In this work, we introduce a real-time, predictive, and task-aware foveated imaging system that operates directly at image acquisition time. Leveraging emerging dual-stream sensor architectures, our method dynamically allocates limited pixel bandwidth to task-relevant regions of interest while maintaining a low-resolution global context. We formulate foveated acquisition as a sensor attention policy--learning problem, in which past observations guide actions that determine future measurements, closing the perception--acquisition loop. Through extensive simulation across multiple perception tasks, we demonstrate that our approach achieves high task performance under strict pixel budgets and significantly outperforms relevant baselines operating at the same bandwidth. We further validate our system on a 200-megapixel dual-stream sensor, capturing real-world videos under realistic bandwidth and latency constraints, demonstrating the practical feasibility of task-driven, acquisition-time foveated imaging. Our project website is at \url{https://howardxiao.ca/foveated/}.

% Ultra-high-resolution image sensors provide rich spatial details critical for modern video perception tasks, but capturing and processing all available pixels is infeasible under real-world latency and power constraints. Existing imaging systems address this challenge through task-agnostic spatiotemporal trade-offs or post-acquisition foveation, both of which discard task-relevant information during capture. In this work, we present a real-time, predictive, and task-aware foveated imaging system that dynamically allocates sensor resolution during acquisition under a limited pixel budget. We formulate foveated image capture as a sensor attention policy learning problem, where past observations are used to predict task-specific regions of interest for future frames. Our lightweight policy operates at acquisition time and closes the perception--acquisition loop. Extensive simulation results demonstrate that our approach maintains full-resolution performance and significantly outperforms conventional methods in pixel-limited settings across multiple video perception tasks. We further validate our system on a 200~MP sensor prototype, demonstrating practical feasibility under realistic bandwidth and latency constraints.
\end{abstract}

\begin{CCSXML}
<ccs2012>
   <concept>
       <concept_id>10010147.10010178.10010224.10010226</concept_id>
       <concept_desc>Computing methodologies~Image and video acquisition</concept_desc>
       <concept_significance>500</concept_significance>
       </concept>
 </ccs2012>
\end{CCSXML}

\ccsdesc[500]{Computing methodologies~Image and video acquisition}
\keywords{computational photography, computational imaging, foveated imaging}

\begin{teaserfigure}
  \includegraphics[width=\textwidth]{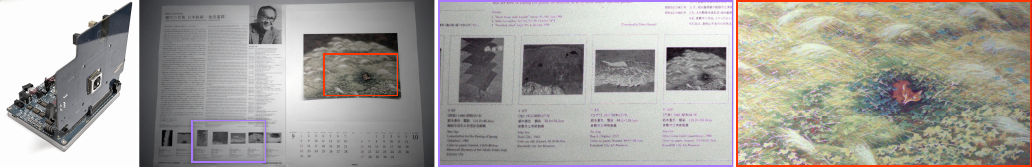}
  \caption{Emerging sensors, such as our prototype based on Samsung's ISOCELL platform (left), provide hundreds of megapixels of resolution, which then fuel modern perception models in imaging, vision, and robotics applications. The massive amount of raw pixel data, however, is extremely challenging to be read out at high frame rates or processed efficiently with downstream perception networks. To address this challenge, we develop a policy-based foveated imaging framework that operates in real time and decides where and when to sample full-resolution regions of interest (center right and right) in a task-specific manner based on low-resolution full-field-of-view context frames (center left) as well as past observations. }
  \label{fig:teaser}
\end{teaserfigure}

\maketitle

\section{Introduction}

Recent advances in image sensor technology enable the capture of ultra-high-resolution images and videos. Commercial sensors beyond 200 megapixels (MP) are widely available~\cite{choi2023world}, with 400 MP prototypes in development~\cite{Canon2025}. Achieving such resolution requires pixel sizes below \(0.6\,\mu\text{m}\) and significantly increases readout bandwidth and downstream processing demands. As a result, under realistic bandwidth, latency, or power constraints, imaging systems cannot afford to acquire, transmit, or process all pixels at full resolution, making selective acquisition essential.

Ultra-high-resolution imagery is increasingly important not only for visual quality, but also for downstream video perception tasks such as object tracking, text recognition, and robotic manipulation. These applications often rely on subtle visual cues---fast-moving objects, small text, or fine-grained surface textures---that are lost at lower resolutions. At the same time, the cost of acquiring and processing ultra-high-resolution video grows prohibitively large as resolution scales. Bandwidth limitations in sensor interfaces, sensor readout, and memory access, along with the quadratic scaling of modern transformer-based perception models with input resolution, further exacerbate this challenge, particularly on edge devices such as augmented-reality glasses, drones, autonomous vehicles, and biomedical systems.

\begin{figure}[!t]
  \centering
   \includegraphics[width=\linewidth]{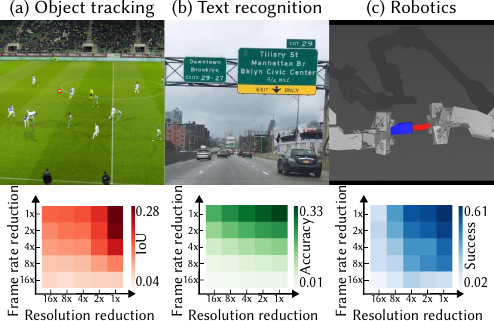}
   \caption{\textbf{Spatio-temporal bandwidth trade-off in different tasks.} Each column shows an example perception task that benefits from both spatial and temporal detail. The top row depicts example inputs, and the bottom row illustrates how task performance varies with spatial and temporal resolution. Task performance is measured by \textbf{(a)} intersection-over-union (IoU), \textbf{(b)} correct transcription rate (Accuracy), and \textbf{(c)} task success rate (Success), consistent with Table~\ref{tab:foveated_imaging}.}
   \label{fig:tradeoff}
\end{figure}

This gap between sensing capability and system constraints raises a fundamental question: \emph{which pixels should be acquired, and when}? Existing systems address this challenge through coarse, task-agnostic spatio-temporal trade-offs, sacrificing either spatial detail or temporal fidelity. Sensors either downsample spatially---through pixel binning or subsampling---to maintain high frame rates, or reduce temporal resolution to preserve spatial detail. While effective at limiting data transmission, these strategies indiscriminately discard high-frequency information that may be critical for perception. Fig.~\ref{fig:tradeoff} illustrates this issue for three different downstream applications. Once lost during acquisition, this information cannot be recovered by subsequent processing, often resulting in degraded performance on detail-critical tasks.

Emerging dual-stream sensors with hundreds of millions of pixels support multiple streams of video data to be read out simultaneously, including low-resolution full-field-of-view frames as well as much smaller but full-resolution regions of interest (ROIs) with a dynamically programmable location in the image~\cite{SamsungISOCELLHP2}. Leveraging these emerging hardware capabilities, in this work we develop real-time, predictive, and task-aware fovea\-tion algorithms that address the problem of determining which pixels to acquire when, under real-world constraints. For this purpose, we formulate foveated image acquisition as a sensor attention policy-learning problem, in which past observations guide actions that directly shape future measurements, closing the perception--acquisition loop for bandwidth-efficient, task-optimal sensing. Our system uses a lightweight saliency module to propose ROI candidates and a task-driven policy to guide ROI evolution during readout. These real-time decisions minimize acquisition bandwidth while preserving task accuracy. Modeling acquisition as sequential decision making enables adaptive, task-driven scanpaths responsive to changes in both the scene and the task objective.

%This paper proposes a real-time, predictive, and task-aware fovea\-ted imaging system that addresses the problem of determining which pixels to acquire when, under real-world constraints. Leveraging emerging dual-stream sensor architectures~\cite{SamsungISOCELLHP2}---already motivated by the need to reduce readout bandwidth and power consumption---our system allocates limited pixel bandwidth to task-relevant regions of interest (ROIs) during sensor readout while maintaining a low-resolution global context. We formulate foveated image acquisition as a sensor-attention-policy-learning problem, in which past observations guide actions that directly shape future measurements, closing the perception--acquisition loop for bandwidth-efficient, task-optimal sensing.

Our approach is motivated by human vision, which leverages the ec\-cen\-tric\-i\-ty-dependent acuity of the retina and eye movements for bandwidth-efficient sensing. Human vision dynamically moves our gaze to fixate the fovea on the most task-relevant regions of a scene. Candidate regions for fixation are typically called salient and a specific sequence of fixations or eye movements is referred to as a scanpath. Scanpaths can exhibit different characteristics, such as saccading or smooth pursuit behavior, again depending on the type of content or the task at hand~\cite{leigh2015neurology}. 

%Emerging dual-stream sensors support multiple streams of content to be read out simultaneously, including low-resolution full-field-of-view context frames as well as a much smaller but full-resolution ROI. Our system uses a lightweight saliency module to propose ROI candidates and a task-driven policy to guide ROI evolution during readout. These real-time decisions minimize acquisition bandwidth while preserving task accuracy. Modeling acquisition as sequential decision-making enables adaptive, task-driven scanpaths responsive to changes in both the scene and the task objective.

Our work makes the following contributions:
\begin{itemize}
    \item We introduce a real-time, policy-based, predictive foveated imaging system that dynamically directs sensor attention during image acquisition.
    \item We demonstrate through extensive simulation that our fo\-ve\-a\-tion approach maintains high task performance and significantly outperforms conventional methods in pixel-limited settings across multiple perception tasks.
    \item We prototype our system using a 200~MP image sensor and capture real-world videos under realistic bandwidth and latency constraints, demonstrating practical feasibility.
\end{itemize}

\section{Related Work}
\label{sec:related}
\begin{figure*}[!t]
  \centering
   \includegraphics[width=.96\linewidth]{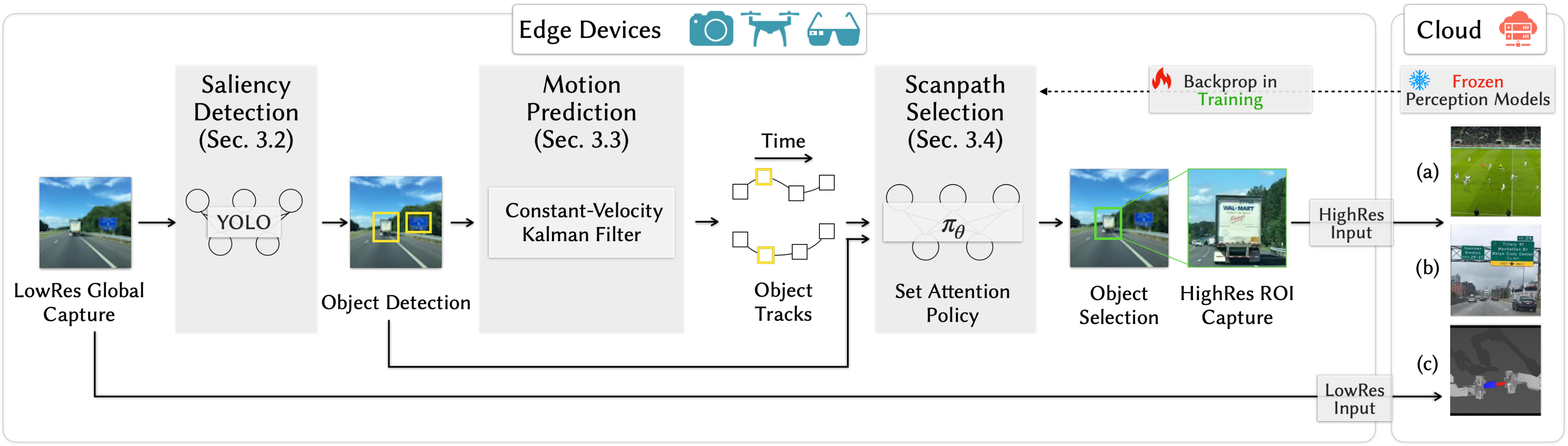}
   \caption{\textbf{Policy-based foveated perception pipeline}. A captured low-resolution frame provides the full-field-of-view global context and is processed to determine salient candidate regions (left). Past observations then guide a per-candidate motion predictor (center left). Our sensor attention policy selects the ROI, which is then read out at full sensor resolution. Both low-resolution context frame and high-resolution ROI are streamed off the edge device and processed by the downstream perception model. At training time, our policy parameters are learned end-to-end with frozen task-specific perception models.}
   \label{fig:method}
\end{figure*}

\subsection{Foveated Computer Vision}
Foveated vision studies how spatial resolution can be allocated non-uniformly across the visual field in order to prioritize task-relevant regions. Early approaches relied on task-agnostic heuristics or saliency cues to identify regions of interest (ROIs) for higher-resolution processing~\cite{karpathy2014large, remmelzwaal2020human, gomes2010towards, itti2002model}. While such methods approximate aspects of human visual attention, they are not optimized for specific downstream perception objectives.

To incorporate task dependence, a large body of work has explored end-to-end learning of foveated representations jointly with perception tasks~\cite{killick2023foveation, killick2025phd, akbas2017object}. Policy-based Recurrent Attention Models (RAM)~\cite{mnih2014recurrent, haque2016recurrent} formulate foveation as a sequential decision-making problem, selecting spatial glimpses conditioned on past observations. More recent works extend this paradigm to video by learning policies that select task-relevant regions from full-resolution inputs for efficient downstream processing~\cite{wang2025emulating, shi2026autogaze}.

Instead of spatial selection, related approaches also address bandwidth or efficiency constraints by temporally subsampling or selectively processing frames~\cite{han2022dynamic, xia2022nsnet}.

Our work is closely related in spirit to prior foveated vision approaches, which all post-process high-resolution image and video data, but it differs in a fundamental assumption: because full-resolution frames cannot be efficiently read out from and transferred off the sensor under real-world bandwidth constraints, our method performs foveation at acquisition time, directly determining which measurements are captured.

\subsection{Active Vision}
Active vision studies how sensing actions can be chosen to improve perception, originally framing sensing as a means to resolve ambiguity and reduce uncertainty~\cite{bajcsy1988active}. Subsequent work explored information-driven and decision-driven viewpoint selection, including next-best-view methods for scene understanding and 3D reconstruction~\cite{connolly1985determination, maver2002occlusions, denzler2002information}, as well as active SLAM systems that plan camera or robot motion to gather informative observations~\cite{sim2005global}.

More recently, learning-based active vision has emerged in embodied perception and neural 3D reconstruction, where sensing actions are optimized for downstream objectives such as robotics policies or reconstruction quality~\cite{kerr2025eye, Chaplot2020Learning, chen2021mvsnerf, chuang2025look}.

Our approach aligns with active vision in closing the perception--action loop, but operates at a different control level: instead of selecting camera poses or viewpoints, our method actively determines the parameters of the camera’s foveation mechanism during runtime.

%we  allocate spatial resolution across the sensor during image acquisition.

\subsection{Foveated Graphics}
Foveation has been widely studied in graphics as a principled way to exploit eccentricity-dependent properties of human vision in order to reduce computation, bandwidth, or power consumption. In foveated rendering and display systems, perceptual models guide level-of-detail and sampling decisions to allocate resources preferentially near the viewer’s gaze fixation~\cite{wang2023foveated, mohanto2021integrative, krajancich2023towards, deng2022fovnerf}.

Although these graphics systems motivate foveation as a principled trade-off between fidelity and efficiency, they typically operate at rendering or display time; in contrast, our method applies foveation during image acquisition, affecting which data is captured rather than post-processing it. We use ``foveated'' to refer to spatially selective, variable-resolution acquisition, generalizing the gaze-contingent interpretation common in graphics.

\subsection{Foveated Sensors}
Spatially-varying resolution has also been realized at the sensor level through multi-aperture and wide-angle lens designs~\cite{carles2016multiaperture, kuniyoshi1995foveated} and event-based sensors~\cite{serrano2022electronically}. The former provide fixed foveation profiles determined by the optics, while the latter produce an output modality that differs from what downstream perception models expect. Our work targets dual-stream image sensors with programmable ROIs, and develops a sensor attention policy that dynamically allocates high resolution during acquisition.

\section{Method}
\subsection{Problem Formulation}

We consider video perception under a strict pixel throughput budget, where an image sensor must dynamically decide \emph{where} and \emph{at what resolution} to acquire visual information in order to maximize downstream task performance.
Assume that the full-resolution video is $\mathbf{v}$ with $\mathbf{v}^{(k)}$ denoting frame $k$, then the sensor observation at frame $k$, $\mathbf{o}^{(k)}$, can be defined as:
\begin{equation}
    \label{eq:obs}
    \mathbf{o}^{(k)} = \mathcal{D}_{\bm{\phi}^{(k)}}(\mathcal{C}_{\bm{\psi}^{(k)}} (\mathbf{v}^{(\mathcal{S}_{\bm{\varphi}}(k))}))
\end{equation}
where we define $\mathcal{D}$ as the spatial downsampling operator with parameters $\bm{\phi}^{(k)} = \{{s_x}^{(k)}, s_y^{(k)}\}$, where $0 < s_x^{(k)}, s_y^{(k)} \leq 1$ represent the spatial pixel resolution reduction factors in $x$ and $y$ directions. Assuming rectangular crops, we denote $\mathcal{C}$ as the frame cropping operator with parameters $\bm{\psi}^{(k)} = \{x^{(k)}, y^{(k)}, w^{(k)}, h^{(k)}\}$ with the top-left corner $(x^{(k)}, y^{(k)})$, cropping width $w^{(k)}$, and cropping height $h^{(k)}$ in pixel space. We further define $\mathcal{S}$ as the temporal skipping operator with parameters $\bm{\varphi} = \{t_s, t_o\}$, where $t_s \in \mathbb{N}$, $t_s \geq 1$ represents the frame skipping stride and $t_o \in \mathbb{N}, t_o \geq 1$ represents the frame offset. Here $t_s, t_o$ are independent of the frame index $k$ and $\mathcal{S}_{\bm{\varphi}}(k) = t_s \cdot k + t_o$ for each $k$.

In this case, $\mathbf{o}^{(k)}$ is parameterized by sensor attention variables $\mathbf{a}^{(k)} = \{\bm{\phi}^{(k)}, \bm{\psi}^{(k)}, \bm{\varphi}\}$ as defined in Eq.~\eqref{eq:obs}.
Given an observation horizon of the past $T_o$ frames, our goal is to predict a sequence of future sensor attentions over a prediction horizon $T_p$:
\begin{equation}
\label{eq:policy_goal}
\mathbf{a}^{(k:k+T_p)} = \pi_\theta \big( \mathbf{o}^{(k-T_o:k)}, \mathbf{c} \big),
\end{equation}
where $\pi_\theta$ denotes a task-conditioned sensor attention policy with parameters $\theta$, and $\mathbf{c}$ encodes optional task-specific conditioning, such as language instructions or visual prompts.
The predicted actions directly determine future observations, closing the perception--acquisition loop. In our setting, dual-stream sensors capture at each frame $k$ a low-resolution global context frame $\mathbf{o}_g^{(k)}$ with fixed downsampling $\bm{\phi}_g^{(k)}<\bm{1}$ over the full frame, and a full-resolution ROI crop with $\bm{\phi}^{(k)}=\bm{1}$ and dynamic crop parameters $\bm{\psi}^{(k)}$. Therefore, the sensor attention reduces to $\mathbf{a}^{(k)}=\{\bm{\psi}^{(k)}\}$.

Rather than learning $\pi_\theta$ end-to-end from raw pixels, we decompose the problem into three lightweight, interpretable components: (i) a saliency detector, (ii) a motion model, and (iii) a scanpath selection policy (Fig.~\ref{fig:method}).
This modular design enables real-time inference on edge hardware and avoids the instability and latency of monolithic policies.

\subsection{Saliency Detection from Low-Resolution Context}
\label{subsec:saliency_detection}
We employ a fast, YOLO-style~\cite{redmon2016you} saliency detector fine-tuned for each downstream task, operating only on context frames $\mathbf{o}_g^{(k)}$. This architecture is chosen for its favorable accuracy--latency trade-off.
The detector outputs a set of $M^{(k)}$ object hypotheses:
\begin{equation}
\label{eq:detections}
\mathcal{B}^{(k)} = \{ (\mathbf{b}_i^{(k)}, \mathbf{f}_i^{(k)}, \ell_i^{(k)}, c_i^{(k)}) \}_{i=1}^{M^{(k)}},
\end{equation}
where $\mathbf{b}_i^{(k)} = (x_i^{(k)}, y_i^{(k)}, w_i^{(k)}, h_i^{(k)})$ is a bounding box in image coordinates,
$\mathbf{f}_i^{(k)}$ is a learned object appearance embedding, $\ell_i^{(k)}$ is the predicted class label, and $c_i^{(k)}$ is the detection confidence score.
Operating exclusively on low-resolution frames ensures minimal acquisition and compute overhead. Using a detector optimized for real-time multi-object localization allows us to efficiently extract global scene structure and object hypotheses under strict runtime constraints.

\subsection{Motion Prediction}
\label{subsec:motion}

To anticipate future object locations at acquisition time, we associate detections across past global frames using the Hungarian matching algorithm~\cite{kuhn1955hungarian} and estimate object motion. This technique is commonly used in multi-object tracking-by-detection algorithms such as SORT~\cite{bewley2016simple} and ByteTrack~\cite{zhang2022bytetrack} and is favored for its real-time performance.
Although motion can be highly non-linear over long horizons in some video perception tasks such as object tracking, our setting requires only short-horizon prediction with frequent receding-horizon replanning, so a simple constant-velocity model provides a sufficiently accurate and low-latency approximation.

For each detected object $i$, we maintain a state vector $\mathbf{s}_i^{(k)}$ consisting of its bounding box center and velocity. We use a constant-velocity Kalman Filter to propagate this state forward:
\begin{equation}
\label{eq:motion}
\hat{\mathbf{b}}_i^{(k+\tau)} = \mathcal{T}_\text{KF} \big( \mathbf{s}_i^{(k)}, \tau \big), \quad \tau = 1, \ldots, T_p,
\end{equation}
yielding predicted bounding boxes for the next $T_p$ frames, which provide the candidate crop parameters $\bm{\psi}^{(k)}$ in Eq.~\eqref{eq:obs}.
This explicit motion model enables low-cost temporal extrapolation and allows the scanpath selection policy to reason over predicted object trajectories while relying on frequent replanning to adapt to rapid motion, occlusions, and interaction dynamics.

\subsection{Scanpath Selection Policy}
\label{subsec:scanpath}

The scanpath selection policy predicts \emph{which objects to foveate and when}.
Rather than predicting continuous ROI parameters directly, the policy outputs a discrete scanpath over detected objects from the saliency detector, which is later converted into ROI parameters $\bm{\psi}^{(k)}$.

\paragraph{Object tokens.}
For each object $i$, we construct a token $\mathbf{z}_i^{(k)}$ by concatenating three components:
\begin{equation}
\label{eq:token}
\mathbf{z}_i^{(k)} =
\big[
\underbrace{\mathbf{r}_i^{(k)}}_{\text{ROI features}} \;\|\;
\underbrace{\mathbf{g}_i^{(k)}}_{\text{global features}} \;\|\;
\underbrace{\mathbf{d}_i^{(k)}}_{\text{detection \& motion features}}
\big].
\end{equation}
Here, $\mathbf{r}_i^{(k)}$ encodes the high-resolution visual features of each past-foveated object $i$ using a frozen MobileNetV3-Small visual encoder backbone specifically optimized for edge device performance~\cite{howard2019searching}. We employ a separate ROI feature encoder because YOLO-style detectors are not trained to extract fine-grained, high-resolution appearance features suitable for general video perception, particularly for small or texture-sensitive objects.
$\mathbf{g}_i^{(k)}$ aggregates low-resolution context features for each object $i$ over the past $T_o$ frames using a temporal 1D convolution network, capturing coarse scene and object context directly from the detector outputs. Finally, $\mathbf{d}_i^{(k)}$ encodes the past bounding box detections, class labels, visibility history, and predicted future boxes $\{\hat{\mathbf{b}}_i^{(k+\tau)}\}_{\tau=1}^{T_p}$ of object $i$.

\paragraph{Global reasoning and prediction.}
Given the set of object tokens $\{\mathbf{z}_i^{(k)}\}$ and an optional task conditioning token $\mathbf{c}$,
we employ a Set Transformer encoder~\cite{lee2019set} to perform permutation-invariant global object reasoning:
\begin{equation}
\label{eq:set_transformer}
\{\tilde{\mathbf{z}}_i^{(k)}\} = \mathrm{SetTransformer}\big( \{\mathbf{z}_i^{(k)}\} \cup \{\mathbf{c}\} \big).
\end{equation}
Each transformed object token is passed through a lightweight multilayer perceptron (MLP) head to predict object selection logits over the next $T_p$ frames.
The logits are then normalized to output a foveation scanpath represented by a categorical distribution over objects at each future timestep $k + \tau$:
\begin{equation}
\label{eq:policy_output}
p_\theta(i \mid k+\tau) = \mathrm{softmax}\big( \mathrm{MLP}(\tilde{\mathbf{z}}_i^{(k)}) \big), \quad \tau = 1,\ldots,T_p.
\end{equation}
The selected object index is mapped to ROI parameters $\bm{\psi}^{(k+\tau)}$ using the corresponding predicted bounding box at each timestep. It is important to note that this formulation naturally incorporates receding-horizon control~\cite{mayne1988receding} that allows the execution of our foveation policy for $T_a < T_p$ future actions before replanning, balancing inference latency and adaptability to changing environments.

\subsection{Why a Modular Policy?}

Our design deliberately separates detection, motion prediction, and foveation scanpath selection.
Compared to a possible end-to-end sensor attention policy, our decomposition offers three important advantages:
(i) real-time inference with predictable latency,
(ii) improved stability and interpretability from component-wise training~\cite{le2018hierarchical}, and
(iii) the ability to swap or upgrade components independently guided by downstream perception tasks.
In practice, the full pipeline runs in real time on CPUs and low-end GPUs with receding-horizon control. It could enable acquisition-time deployment on edge devices. We provide a detailed runtime analysis in the supplemental material.
% In practice, the full pipeline runs in real time on CPUs and low-end GPUs with receding-horizon control. It could enable acquisition-time deployment on edge devices. We provide a detailed runtime analysis in the supplemental material.
% \subsection{Unsupervised Training of Scanpath Selection Policy}
% Not needed anymore (jan)

\section{Evaluation and Experiments}
\label{sec:results}
\begin{table*}[!ht]
    \caption{\textbf{Quantitative results of policy-based foveated perception.} We compare downstream task performance of our method against same-pixel-bandwidth downsampling baselines and include full-resolution and oracle performance with ground truth (GT) ROI selections. \textbf{Row 1:} We compare downstream soccer ball tracking Intersection-over-Union (IoU) of the baseline methods and our approach. Temporal downsampling IoU is computed only on kept frames. \textbf{Row 2:} We compare the percentage of distinct text objects correctly detected and transcribed in the road scene text recognition task. \textbf{Row 3:} We compare the partial and complete task success rate over 100 trials of the ALOHA insertion task. The GT Oracle is not applicable because no ground-truth foveation labels are available for this task. Our approach performs the best among relevant baselines with a comparable bandwidth.}
    \label{tab:foveated_imaging}
    \centering
    \begin{tabular}{c c c | c | c c c}
        \toprule
        Task & Metric & \makecell[c]{Full-resolution} & GT Oracle & \makecell[c]{Spatial \\ downsampling} & \makecell[c]{Temporal \\ downsampling} & \makecell[c]{\textbf{Foveated} \\ \textbf{(Ours)}} \\
        \midrule
        Object Tracking & IoU $\uparrow$ & 0.281 & 0.405 & 0.122 & 0.148 & \textbf{0.283}\\
        \midrule
        Text Recognition & Transcription Rate $\uparrow$ & 0.333 & 0.271 & 0.067 & 0.248 & \textbf{0.264}\\
        \midrule
        Robotic Manipulation & \makecell[c]{Success Rate $\uparrow$ \\ (Complete $\mid$ Partial)} & 0.15 $\mid$ 0.61 & N/A & 0.10 $\mid$ 0.51 & 0.07 $\mid$ 0.30 & \textbf{0.12 $\mid$ 0.57} \\
        \bottomrule
    \end{tabular}
\end{table*}

\begin{figure*}[!t]
    \centering
\includegraphics[width=\linewidth]{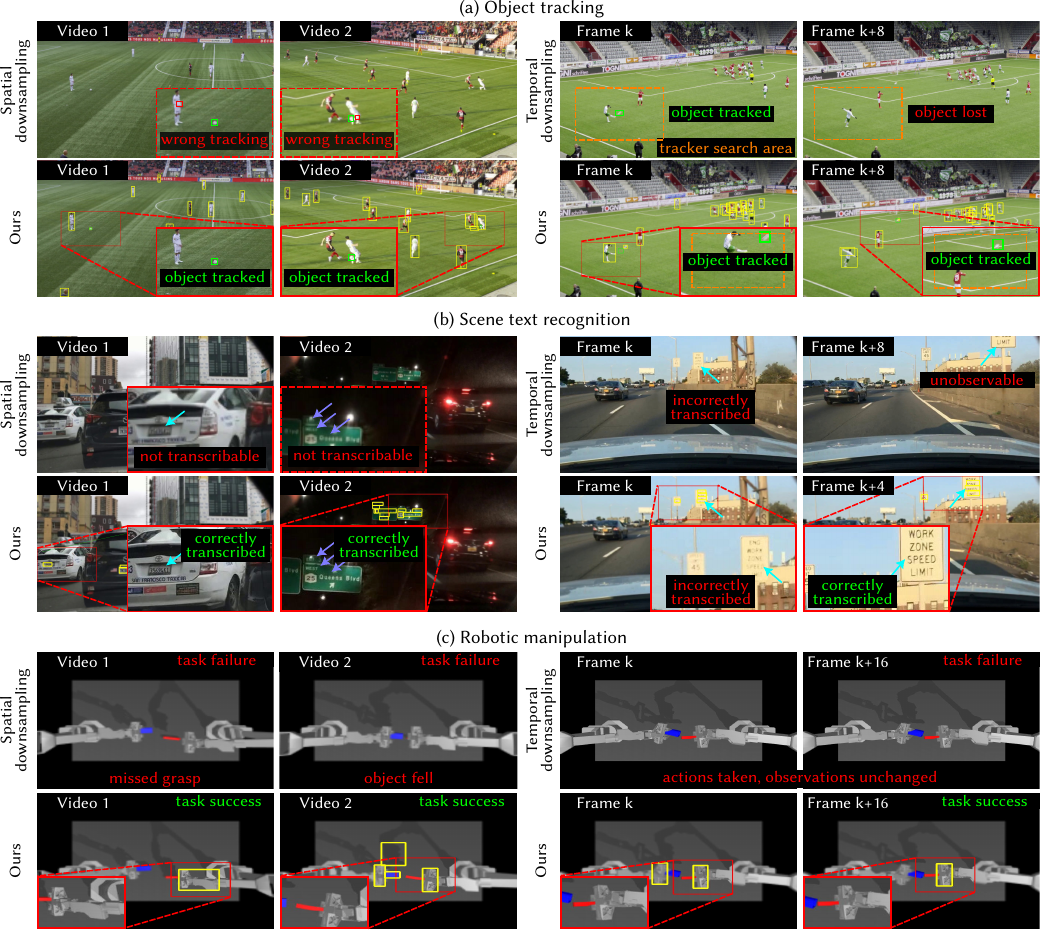}
    \caption{\textbf{Policy-based foveated imaging and perception for simulated video tasks.} \textbf{Row (a):} Our foveated imaging approach correctly allocates higher resolution for pursuing objects of interest in an object tracking task. ROIs from our foveated imaging framework provide fine spatial details required to distinguish similar objects and provide fine temporal details for motion continuity, significantly improving downstream search-based tracker performance compared to task-agnostic spatio-temporal downsampling baselines. \textbf{Row (b):} Our method adapts to emerging objects and allocates fine resolution to high-frequency text regions important for the downstream scene text recognition task before frames are captured. Our foveated imaging pipeline improves the transcription rate of the text recognition task compared to naive spatio-temporal downsampling methods where texts are frequently not transcribable or missed. \textbf{Row (c):} In robotic manipulation, our method attends to important regions critical for task success while keeping low latency, ensuring the observed state reflects robot actions and significantly improving our performance against task-agnostic baselines. Additional results across more diverse scenes are provided in the supplemental material.}
    \label{fig:results_qual}
\end{figure*}

We evaluate our foveated imaging framework on multiple video perception tasks in simulation. Our experiments are designed to answer three questions: (1) can the policy predict task-relevant regions of interest (ROIs) \emph{before} high-resolution measurements are captured; (2) does policy-based foveated imaging improve downstream video perception under strict pixel bandwidth constraints compared to task-agnostic acquisition strategies; and (3) can such a predictive foveated imaging system be realized in practice on an ultra-high-resolution imaging platform operating under realistic latency and bandwidth limits.

Unless otherwise specified, all downstream perception models are kept frozen during evaluation to isolate the effect of the acquisition strategy. This design demonstrates that our foveated imaging framework can be layered on top of existing perception models, minimizing the need for fine-tuning or post-training.

\subsection{Experimental Protocol}

All methods are evaluated under explicitly controlled pixel bandwidth constraints. For a given budget, we ensure that the average pixel throughput over time is identical across all acquisition strategies, including spatial downsampling, temporal downsampling, and our policy-based foveated imaging method.
The pixel budgets relative to full resolution are $1/8$ for object tracking, $1/8$ for scene text recognition, and $1/16$ for robotic manipulation; additional results at other pixel budgets are provided in the supplemental material.
The same downstream model, dataset split, and evaluation metric are used across acquisition strategies for each task. Additional implementation details, including policy architecture, training procedures, and hyperparameters, are provided in the supplemental material.
\subsection{Tasks, Models, and Metrics}

We evaluate three video perception tasks
% that place
with
% complementary
different
demands on spatial detail, temporal resolution, and closed-loop responsiveness.

For object tracking, we use the SoccerNet Tracking dataset~\cite{cioppa2022soccernet}, which features 1920$\times$1080 high-resolution video clips with fast-moving targets, large camera motion, and frequent occlusions. We use MixFormerV2~\cite{cui2023mixformerv2} as the downstream tracker, which outputs a bounding box per frame. Performance is measured using Intersection over Union (IoU) against ground-truth annotations. We evaluate three tracking subjects: the soccer ball, referees, and players.

For scene text recognition, we evaluate on the RoadText-1K dataset \cite{reddy2020roadtext}, which contains 1280$\times$720 outdoor road-scene videos with small and sparsely distributed text regions. We use DeepSolo~\cite{ye2023deepsolo} as the downstream model, which performs joint text detection and transcription. Performance is measured using the end-to-end correct transcription rate.

For robotic manipulation, we evaluate on the Static ALOHA dataset~\cite{zhao2023learning}, which consists of tabletop manipulation tasks that are highly sensitive to spatial detail and temporal feedback. Experiments are conducted in simulation, with frames rendered at 640$\times$480 resolution. We use the pretrained task-specific ALOHA Action Chunking Transformer (ACT)~\cite{zhao2023learning} as the downstream model, which predicts action chunks executed by a receding-horizon controller that replans every 15 steps. Following the original benchmark definition, performance is measured by partial and complete task success rates, where partial success corresponds to achieving stable contact between the manipulated objects, and complete success requires correctly inserting one object into the other.

% Together, these tasks span frame-wise perception, structured recognition, and embodied control, providing a broad testbed for evaluating task-aware foveated imaging.
\subsection{Acquisition Baselines}
\label{subsec:baselines}

We compare our approach against task-agnostic acquisition strategies operating under the same pixel budget. Spatial downsampling uniformly reduces the spatial resolution of the full frame while preserving the original frame rate, trading spatial detail for temporal smoothness. Temporal downsampling reduces the frame rate while maintaining full spatial resolution, preserving fine details at the cost of temporal continuity.
Both baselines represent common approaches used in current ultra-high-resolution sensors for video acquisition and perception. They allocate pixels uniformly and do not adapt acquisition decisions based on scene dynamics or task objectives, allowing us to isolate the benefits of our policy-based, task-guided foveated imaging approach. We further include a \emph{GT Oracle} upper bound that bypasses all components in Secs.~\ref{subsec:saliency_detection}--\ref{subsec:scanpath} and directly centers ROIs on the target’s ground-truth bounding boxes under the same pixel budget.

\subsection{Downstream Video Perception under Limited Pixel Budget}
\begin{figure*}[!t]
    \centering
    \includegraphics[width=\linewidth]{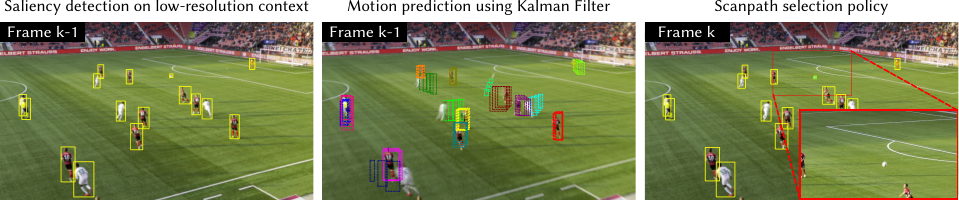}
    \caption{\textbf{Overview of key components of our foveated imaging framework.} \textbf{Left:} At frame $k - 1$, multiple task-relevant object locations are proposed by the saliency detector using only low-resolution context. \textbf{Middle:} We associate each object detection over the past $T_o$ frames and predict object motion for the future $T_p$ frames using a constant-velocity Kalman Filter. \textbf{Right:} Our scanpath selection policy uses high-resolution ROI features, global features, and detection and motion features to predict the future object scanpath at frame $k$ from detected objects in the past. The predicted scanpath allows us to acquire future ROIs with high resolution for downstream perception tasks.}
    \label{fig:qual_pipeline}
\end{figure*}

We evaluate whether predictive foveated imaging improves downstream task performance under a limited pixel budget compared to task-agnostic baselines in Sec.~\ref{subsec:baselines}. For each task, we compare (i) full-resolution inputs, (ii) dual-stream inputs acquired using predicted high-resolution ROIs with downsampled global context (Foveated), (iii) spatially downsampled inputs, and (iv) temporally downsampled inputs, with (ii), (iii), and (iv) matched to the same total pixel bandwidth. Downstream models are fixed per task.
Table~\ref{tab:foveated_imaging} summarizes the results. Overall, policy-based foveated imaging consistently outperforms task-agnostic baselines and, in some cases, matches full-resolution performance while using less than one-eighth of the pixel bandwidth.

\paragraph{Object tracking.} Objects in SoccerNet Tracking are small and fast-moving, making tracking particularly sensitive to acquisition bandwidth. The soccer ball occupies only a few pixels on average (approximately $15$ at full resolution) and, after naive spatial downsampling, falls well below the effective patch size of downstream transformer-based models, leading to severely degraded localization accuracy (IoU $0.122$). Temporal downsampling performs slightly better (IoU $0.148$), but fails under fast motion: the ball often traverses more than $5\%$ of the field of view in less than $10$ frames, making it difficult for search-template-based trackers to reliably establish correspondences (see Figs.~\ref{fig:results_qual} and~\ref{fig:qual_pipeline}).

In contrast, our predictive foveated imaging framework tracks the ball’s trajectory despite rapid motion, achieving an IoU of $0.283$ while operating at $8\times$ lower bandwidth and effectively matching full-resolution performance (IoU $0.281$). Our method and the GT Oracle both outperform the full-resolution baseline, as passing ROI crops suppresses background distractors and improves tracking despite using fewer pixels, consistent with findings in~\cite{zhu2018distractor}.

Beyond the soccer ball, the policy adapts its foveation behavior online by changing visual conditioning, enabling smooth pursuit of different objects---including players and referees---within the same video. Its causal, acquisition-time operation allows rapid adaptation to occlusions, abrupt motion, and potential identity switches.

\paragraph{Text recognition.} Text in RoadText-1K appears only briefly and at varying distances as the ego vehicle moves; text on other vehicles or roadside signs may enter and exit the field of view rapidly and can be difficult to read when small or partially occluded. Under these conditions, spatial downsampling leads to a severe drop in transcription accuracy ($0.067$), as fine character strokes become unrecognizable. Temporal downsampling performs better ($0.248$), but remains unreliable because text is often readable only within a narrow temporal window that may be skipped entirely. These failure modes are illustrated in Fig.~\ref{fig:results_qual}, where spatial downsampling blurs text beyond recognition while temporal subsampling skips the few frames in which text might be legible.

Our foveated approach achieves a transcription rate of $0.264$, outperforming both bandwidth-matched baselines by preserving high-resolution detail over text regions while maintaining sufficient temporal coverage. More broadly, RoadText-1K highlights the inherent difficulty of text recognition under bandwidth constraints: multiple text instances may be simultaneously present, and limited pixel budgets require explicit decisions about where to allocate resolution. This is reflected in the gap between full-resolution performance ($0.333$) and the GT Oracle ($0.271$), which is close to our result.

\paragraph{Robotic manipulation.} The Static ALOHA bimanual insertion task requires high dexterity and tight coordination between perception and control, as successful execution depends on precise localization of contact regions and timely visual feedback during closed-loop manipulation. Complete success is more challenging than partial success, as it requires higher precision across all task-relevant dimensions; accordingly, every instance of complete success also constitutes partial success.
% Following the original benchmark definition, we define partial success as achieving stable contact between the two manipulated objects, while full success requires correctly inserting one object into the other.

Temporal downsampling severely degrades partial success (from $0.61$ to $0.30$), as reduced visual feedback causes the controller to overshoot actions without receiving intermediate corrective signals. Spatial downsampling also reduces partial success (to $0.51$), though to a slightly lesser extent, reflecting the loss of fine spatial detail needed for accurate alignment between the robot end-effector and the manipulated objects.

In contrast, our foveated imaging framework preserves high-resolution sensing over task-relevant regions---such as the end-effectors and object interaction points---while maintaining sufficient temporal feedback. As a result, our method achieves performance close to that of full-resolution sensing (Table~\ref{tab:foveated_imaging}). We observe similar trends for complete success.

\subsection{Ablation Studies}

We conduct ablation studies to isolate the contribution of individual components in our foveated imaging framework. All ablations are evaluated on the three tasks described above. To enable comparison across heterogeneous metrics, we normalize each task’s performance relative to its full-resolution performance, preserving relative degradation trends while allowing aggregation across tasks.

\paragraph{System-level ablations.} We compare our learned, task-aware foveation policy with simpler alternatives: always-centered ROI selection and deterministic round-robin ROI scanning. Results in Table~\ref{tab:ablation_system} show that always selecting a centered ROI performs extremely poorly for object tracking and text recognition, as task-relevant content is rarely centered in dynamic scenes. While this strategy performs moderately well for robotic manipulation---where the end-effector often remains near the image center---it fails to generalize across tasks. Round-robin scanning improves over fixed centering by ensuring spatial coverage, but remains substantially inferior to our method, particularly for soccer tracking and manipulation. In contrast, our policy-based approach consistently achieves near- or above-full-resolution normalized performance across all tasks.

\begin{table}[!t]
    \caption{\textbf{System-level ablation results averaged across all tasks.} Performance is normalized by full-resolution performance per task for cross-comparison. Our dynamic foveation approach outperforms naive ROI selection strategies. Normalized performance here can exceed 100\% whenever ROI-based perception outperforms direct processing of the full-resolution image, which contains additional visual clutter in non-task-relevant areas.}
    \label{tab:ablation_system}
    \centering
    \begin{tabular}{l | c c c }
        \toprule
        Variant & \makecell[c]{Soccer \\ Tracking} & \makecell[c]{Text \\ Recognition} & \makecell[c]{Robotic \\ Manipulation}\\
        \midrule
        \makecell[l]{Always-centered \\ ROI selection} & 1.7\% & 3.6\% & 82.0\% \\
        \midrule
        \makecell[l]{Round-robin \\ ROI scanning} & 3.5\% & 64.0\% & 72.1\%\\
        \midrule
        Ours & \textbf{100.7\%} & \textbf{79.3\%} & \textbf{93.4\%}\\
        \bottomrule
    \end{tabular}
\end{table}

\paragraph{Policy input feature ablations.} We further analyze the contribution of individual policy inputs by selectively removing feature groups from the scanpath selection policy, including high-resolution ROI features, low-resolution global context features, and motion features derived from short-term trajectory prediction.

Table~\ref{tab:ablation_features} shows that removing motion features causes a performance drop across all tasks, underscoring the importance of anticipating future object locations when making acquisition-time decisions. Removing global context features also leads to significant degradation, particularly for text recognition, where multiple candidate regions may be present and scene-level context is required to disambiguate where high-resolution sensing should be allocated. Removing high-resolution ROI appearance features also degrades tracking, text recognition, and manipulation performance, reflecting the importance of fine-grained visual detail.

Overall, these ablations confirm that effective acquisition-time foveation requires the combination of motion cues, global context, and high-resolution local appearance. Removing any of these components degrades performance, whereas their integration enables robust task-aware sensing under strict pixel-bandwidth constraints.

\begin{table}[!t]
    \caption{\textbf{Ablation of policy inputs.} Performance is normalized by full-resolution performance per task for cross-comparison. All features are necessary for the best performance; see Eq.~\eqref{eq:token}.}
    \label{tab:ablation_features}
    \centering
    \begin{tabular}{l | c c c }
        \toprule
        Policy Inputs Removed & \makecell[c]{Soccer \\ Tracking} & \makecell[c]{Text \\ Recognition} & \makecell[c]{Robotic \\ Manipulation}\\
        \midrule
        \makecell[l]{w/o ROI features} & 53.3\% & 78.7\% & 83.6\% \\
        \midrule
        \makecell[l]{w/o global features} & 99.2\% & 72.7\% & 85.2\%\\
        \midrule
        \makecell[l]{w/o motion features} & 90.0\% & 71.8\% & 90.2\%\\
        \midrule
        All features (ours) & \textbf{100.7\%} & \textbf{79.3\%} & \textbf{93.4\%}\\
        \bottomrule
    \end{tabular}
\end{table}

\section{Evaluation on a 200~MP Foveated Imaging Prototype}
\label{sec:prototype_eval}
\begin{figure*}[!t]
    \centering
    \includegraphics[width=\linewidth]{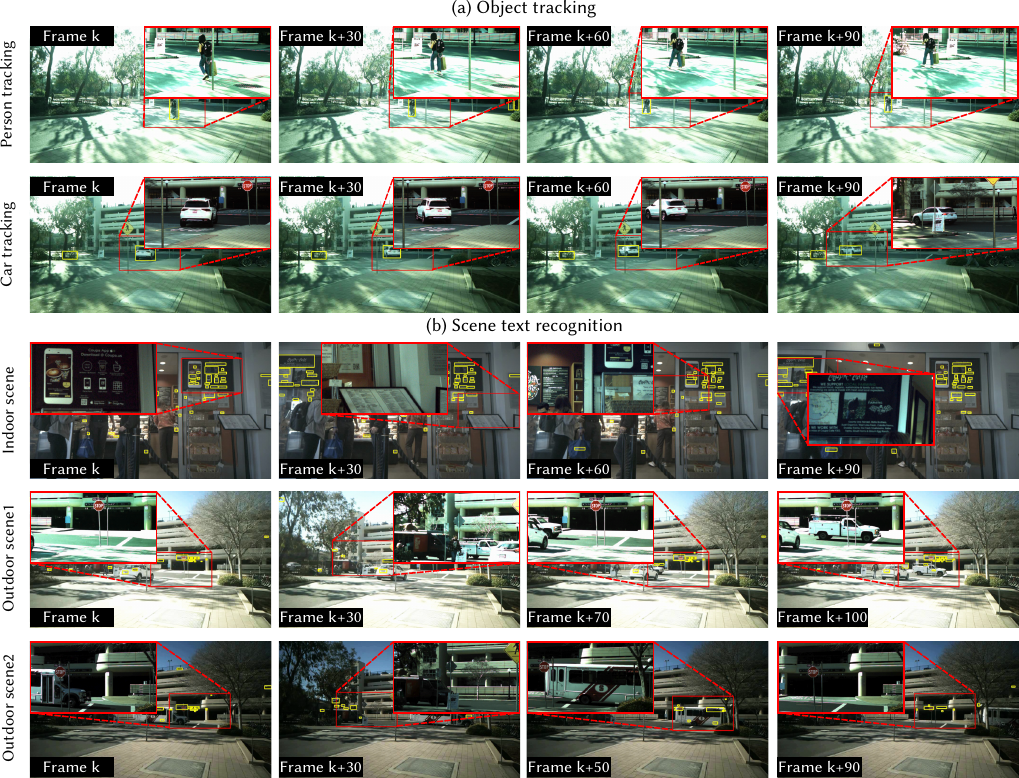}
    \caption{\textbf{Policy-based foveated imaging in real-world captures.} Under realistic bandwidth and acquisition latency constraints, our proposed method runs in real time on our 200~MP-resolution foveated imaging prototype. We demonstrate expected smooth-pursuit scanpaths for \textbf{(a)} object tracking and saccading scanpaths for \textbf{(b)} scene text recognition across diverse scenes and lighting conditions.}
    \label{fig:prototype}
\end{figure*}

To validate the practical feasibility of predictive foveated imaging, 
%we evaluate our framework on a physical 200~MP dual-stream imaging prototype. This evaluation demonstrates that our learned attention policy operates reliably within the strict timing and bandwidth constraints of a real acquisition pipeline, while adapting to real-world dynamics and sensor noise that are difficult to model in simulation.
%
we implement our predictive foveated imaging framework on a hardware prototype built around a 200~MP Samsung ISOCELL HP2 image sensor. The sensor is mounted on a custom control board that supports dual-stream acquisition, enabling simultaneous capture of a low-resolution Full Field-of-View (FFoV) context stream and high-resolution Region-of-Interest (ROI) crops at 30 frames per second. The control board is interfaced with a host system via a Python API, which allows predicted ROI coordinates to be transmitted to the sensor for subsequent frame readout.

In our prototype configuration, the FFoV stream is captured at a resolution of $2040 \times 1148$, providing global situational awareness, while each ROI occupies one-quarter of the sensor area and is captured at $4080 \times 2296$ resolution, corresponding to a $16\times$ increase in spatial resolution relative to the FFoV stream. The resolution difference between the FFoV and the ROI stream matches our simulation setting across all video perception tasks. This dual-stream setup enables closed-loop, predictive control of sensor readout, allowing high-resolution sensing resources to be dynamically allocated to task-relevant regions during acquisition.

\subsection{Hardware and Predictive Acquisition Loop}
\label{subsec:hardware_arch}

The prototype implements a predictive dual-stream acquisition loop consisting of four stages. First, in the Low-Resolution Context Capture stage, the sensor reads out an FFoV context frame at $2040 \times 1148$ resolution. Second, during Policy Inference, the host controller processes the FFoV frame using our lightweight attention policy to predict the Region-of-Interest (ROI) coordinates for the subsequent frames ($k{+}\tau$). Third, in the Command Transmission stage, the predicted ROI coordinates are transmitted back to the sensor controller via a Python-to-FPGA interface. Finally, in the High-Resolution ROI Capture stage, the sensor acquires the targeted ROI at $4080 \times 2296$ resolution, providing $16\times$ higher spatial detail than the FFoV stream.

This predictive loop ensures that ROI selection decisions are made prior to high-resolution readout, allowing the system to manage sensor bandwidth proactively rather than reactively.

\subsection{Real-World Performance and Bandwidth Efficiency}
\label{subsec:real_world_perf}

We capture dual-stream Bayer-raw video at 30~fps, with optics manually focused prior to acquisition. Despite the computational overhead of policy inference and bidirectional host–sensor communication, the system maintains a stable end-to-end throughput of 30~fps throughout extended capture sessions.

Qualitative results are shown in Fig.~\ref{fig:teaser} and Fig.~\ref{fig:prototype}, where we demonstrate both smooth-pursuit scanpaths for object tracking and saccading scanpaths for scene text recognition. The predictive attention policy consistently directs high-resolution sensing to task-relevant regions that remain indistinguishable in the FFoV context stream. This enables recovery of fine spatial details such as object boundaries and textures under real-world lighting conditions and sensor noise.

Crucially, the system achieves this performance while reading out only $6.25\%$ of the sensor area at full resolution per frame. This demonstrates that predictive foveated imaging provides an effective mechanism for managing the bandwidth of 200~MP-class sensors, preserving task-critical visual information while operating within realistic hardware and interface constraints.

\section{Discussion}

%\paragraph{Discussion.}
Our results suggest that predictive, policy-based foveated acquisition is a promising approach for operating ultra-high-resolution image sensors under realistic bandwidth, latency, and power constraints. By explicitly modeling the interaction between sensing and perception, our framework allocates limited pixel budgets to task-relevant regions \emph{before} high-resolution measurements are captured, preserving downstream performance that would otherwise degrade under conventional spatio-temporal downsampling.

\paragraph{Limitations and Future Work.} Despite these advantages, our approach has several limitations. First, the effectiveness of predictive foveation depends on temporal coherence: tasks involving highly stochastic or instantaneous events may reduce the benefit of anticipation. Second, while our attention policy is lightweight, it introduces additional system complexity and must meet strict real-time constraints to be deployed at acquisition time. Third, our current prototype supports a limited number of ROIs per frame; extending the framework to support more flexible or hierarchical foveation patterns remains an interesting direction for future work.

More broadly, our formulation highlights foveated imaging as a systems problem that spans sensor design, learning-based control, and downstream perception models. While we focus on a specific set of video perception tasks, the proposed framework is general and could be extended to other sensing modalities, multi-camera systems, or closed-loop robotic perception pipelines.

%In this work, we demonstrate successful application of our method to several important and diverse applications. Many more could be explored in the future. Our sensor attention policy currently runs in an open-loop setting, implying that the output of the downstream perception model does not influence the sensor policy. In the future, a closed-loop principle could be developed and further optimize the performance. Our current implementation runs efficiently on lower-end GPUs or even CPUs; future work could optimize this code base for edge processors. \gw{comment on limitations and future work of prototype}

\paragraph{Conclusion.}
Intelligent sensing moves beyond passive capture, enabling systems to decide where and how to sample based on the task at hand. Our policy-based foveation framework breaks conventional sampling trade-offs, providing a lightweight yet powerful solution for high-stakes environments. Inspired by the successes of imitation and reinforcement learning, we believe this paradigm will redefine the boundaries of intelligent data acquisition in computer vision, robotics, and beyond.

\begin{acks}
We thank Samsung for their support with the ISOCELL development kit. Howard Xiao is supported by Stanford Graduate Fellowships (SGF). We thank Hansheng Chen, Ryan Po, Kiyohiro Nakayama, and Zichun Xu for fruitful discussions. Compute resources were provided by the Marlowe cluster at the Stanford University~\cite{kapfer2025marlowe}.
\end{acks}

\balance
\bibliographystyle{ACM-Reference-Format}
\bibliography{main}

\clearpage
\appendix
\twocolumn[
\begin{center}
    {\textbf{Supplementary Material for Policy-based Foveated Imaging and Perception}}
    \vspace{1em}
\end{center}
]
\section{Method Details}
\subsection{Sensor Attention Parameters (Main Paper Sec.~3.1)}
Our framework defines pixel bandwidth as a fraction of total pixels and is therefore resolution-agnostic. In our simulation results (Main Paper Sec.~4), we use SoccerNet, RoadText-1K, and ALOHA datasets at their native resolutions for reproducibility, and validate scalability to ultra-high-resolution captures using our 200~megapixel (MP) hardware prototype (Main Paper Sec.~5). The spatiotemporal tradeoffs analyzed in Main Paper Fig.~2 already emerge at these simulation resolutions.

In both the object tracking and text recognition tasks, we use $0.25\times$ downsampled global images from the full-resolution images in the dataset as our policy input; Region-of-Interest (ROI) crops occupy $1/16$ of the entire field of view and maintain aspect ratios, reducing the full-resolution bandwidth by more than $8\times$. In the robotic manipulation task, global images are downsampled by $0.17\times$ ($32\times$ less bandwidth than full-resolution), and ROI crops occupy $1/16$ of the entire field of view at a slight $0.7\times$ downsampling ($2\times$ less bandwidth than full-resolution), reducing the full-resolution bandwidth by more than $16\times$. We do not apply temporal subsampling to the foveated acquisitions in downstream task evaluations across all simulated experiments.

\subsection{Saliency Detection from Low-Resolution Context (Main Paper Sec.~3.2)}
We use a YOLO-style detector across all simulated and real experiments for saliency detection. The YOLO-11~\cite{khanam2024yolov11} Nano model family is chosen for its favorable speed.

In the object tracking task, we randomly partition the SoccerNet Tracking Dataset~\cite{cioppa2022soccernet} training split into 80\% training and 20\% validation. We finetune a YOLO-11 Nano model on the training split for 50 epochs with a batch size of 128. Other hyperparameters, including the input image size, are kept at their default values. We emulate low-resolution global frames by downsampling the training images with the same downsampling parameter ($0.25\times$) for policy training. Bounding boxes from the tracking dataset are labeled as one of the three classes: player, referee, and ball. We select the best-performing model on the validation split and use it for saliency detection across all object tracking simulation experiments. We report precision and recall across all classes for the best-performing model on the validation split in Table~\ref{tab:finetune_yolo}. We use default hyperparameters at inference, including confidence thresholds and Non-Maximum Suppression (NMS) values.

In the text recognition task, we randomly partition the RoadText-1K dataset~\cite{reddy2020roadtext} into 80\% training and 20\% validation, and finetune a YOLO-11 Nano-OBB model that predicts Oriented Bounding Boxes (OBB), as text is often not horizontal. Since RoadText-1K videos contain many frames with repetitive text detections, we further aggregate the training dataset using images and labels from TextOCR~\cite{singh2021textocr}. We finetune the model for 30 epochs with a batch size of 128, keeping the other hyperparameters at their default values. We emulate low-resolution global frames by downsampling the training images with the same downsampling parameter ($0.25\times$) for policy training. Only one prediction class is used, as we are only interested in text predictions. We select the best-performing model on the validation split and use it for saliency detection in all scene text recognition simulation experiments. We report the precision and recall of the best-performing model on the validation split in Table~\ref{tab:finetune_yolo}. The same model is used for saliency detection in our real-world captures discussed in Main Paper Sec.~5.2. We use default hyperparameters at inference, including confidence thresholds and Non-Maximum Suppression (NMS) values.

For the robotic manipulation task, as no labeled saliency bounding boxes are available, we use the pre-trained YOLO-11 Nano model out-of-the-box and set the prediction confidence threshold to 0.02, the NMS value to 0.1, and enable class-agnostic NMS at inference.

Unlike downstream perception tasks (e.g.\ scene text recognition) that require fine-grained high-resolution details, the saliency detector only needs to \emph{localize} candidate objects at the bounding-box level using coarse global features. Operating on low-resolution global frames is therefore sufficient for our saliency module, and our finetuned detectors achieve $0.88/0.78$ precision/recall on tracking and $0.62/0.43$ on scene text recognition at $0.25\times$ downsampled input (Table~\ref{tab:finetune_yolo}). The robotic manipulation task further shows that the module generalizes to unseen environments: even though the ALOHA object classes do not match the pre-trained YOLO-11 Nano model's training labels, its detections are accurate enough for our policy to select task-relevant ROIs.

\subsection{Motion Prediction (Main Paper Sec.~3.3)}
We use the same hyperparameter settings for object association and motion prediction across all simulated and real experiments. For Hungarian matching, we use a minimum Intersection-over-Union (IoU) threshold of 0.1 and prioritize same-class detections. For motion prediction, we use a constant-velocity Kalman Filter with process noise set to $0.001$ and measurement noise set to $0.01$. When a tracklet is missing detections in the past $T_o$ frames, we gap-fill the detections with a Kalman Filter using the same hyperparameters.

\subsection{Scanpath Selection Policy (Main Paper Sec.~3.4)}
We use the same Set Transformer~\cite{lee2019set} architecture for our scanpath selection policy across our simulated and real experiments.

High-resolution ROI features, global features, and detection and motion features are 64-dimensional each. High-resolution ROI object features are first extracted from the MobileNetV3-Small~\cite{howard2019searching} visual encoder for objects contained in the previous foveated region, and then we use a 1D CNN to aggregate across the past $T_o$ frames and output a 64-dimensional feature vector $\mathbf{r}_i^{(k)}$. Global features are directly extracted from the YOLO model's stride-8 layer using the RoIAlign method from Mask R-CNN~\cite{he2017mask} and temporally aggregated with a 1D CNN that uses different weights from the high-resolution feature's 1D CNN. Detection and motion bounding boxes are simply concatenated over time, and each stream is then projected to a $32$-dimensional vector using a multi-layer perceptron (MLP).

Each object token $\mathbf{z}_i^{(k)}$ is a 192-dimensional feature vector. Each component (ROI features, global features, detection features, and motion features) is normalized independently using separate LayerNorm modules before we project the feature vector into a 128-dimensional embedding space. Object tokens are optionally concatenated with a conditioning token and pass through a standard 2-layer Transformer encoder with 4 attention heads and a 256-dimensional feedforward network. The Set Transformer allows permutation-invariant exchange of information about an object's spatial relationships, temporal dynamics, and task objective via the conditioning token. We use a maximum of 25 objects and attention masking across all simulated and real experiments. Only the object tokens are passed through the final MLP head, which produces selection logits over the next $T_p$ frames. We apply softmax to obtain a categorical distribution over objects and sample from it for scanpath selection.

\paragraph{Training details.} For all simulated experiments, during training and inference, we set $T_o = 4$ frames and $T_p = 16$ frames. During inference, we use receding-horizon control with a replanning interval of $T_a = 8$ frames. We use a batch size of $128$ and train with the AdamW optimizer and a learning rate of $10^{-4}$ with cosine learning rate decay. For the object tracking task, we train for $3{,}000$ iterations; for the scene text recognition task, $5{,}000$ iterations; and for the robotic manipulation task, $10{,}000$ iterations. $10\%$ of the iterations are spent on learning rate warmup. For both simulated and real experiments, the total training time for all components of our foveated imaging pipeline is less than 12 hours on a single GPU with about 20~GB of VRAM.

\subsection{Runtime Analysis (Main Paper Sec.~3.5)}
Our policy is designed and optimized for real-time performance on low-end GPUs and CPUs. Table~\ref{tab:runtime} summarizes runtime estimates for the different components of our foveated imaging framework on the laptop CPU used for our real-world captures. Most runtime is spent in the saliency detection module; despite this, our lightweight foveated imaging framework achieves real-time performance on CPUs with receding-horizon control.

The runtimes in Table~\ref{tab:runtime} are measured on an Intel Core i7-7700HQ laptop CPU (16~GB of RAM) that also drives our 200~MP prototype via a Python API over USB 3.0. At 30~fps capture with receding-horizon control at $T_a = 8$ frames, three replans complete within each second, giving an aggregate compute of less than 80~GFLOPs/sec, dominated ($>98\%$) by saliency detection. This compute requirement is well within the budgets of modern NPU-accelerated edge platforms such as smart glasses and drones.

\begin{table}[!t]
    \caption{\textbf{Saliency detection model evaluation}. The finetuned models demonstrate reasonable capability of detecting salient objects in unseen and blurry images.}
    \label{tab:finetune_yolo}
    \centering
    \begin{tabular}{c c c}
        \toprule
        Task & Precision & Recall \\
        \midrule
        Object tracking & 0.88 & 0.78  \\
        \midrule
        Scene text recognition & 0.62 & 0.43 \\
        \bottomrule
    \end{tabular}
\end{table}

\begin{table}[H]
    \caption{\textbf{Foveated imaging pipeline runtime evaluation}. Our full pipeline achieves about 2 frames of latency on a laptop CPU during 30~fps video capture.}
    \label{tab:runtime}
    \centering
    \begin{tabular}{c c c c}
        \toprule
        Component & Parameters & GFLOPs / call & Runtime [ms]\\
        \midrule
        \makecell[c]{YOLO \\ (Saliency detection)} & 2.6M & 6.5 & 55 \\
        \midrule
        \makecell[c]{MobileNetV3 \\ (ROI feature)} & 2.5M & 0.12 & 6.2 \\
        \midrule
        \makecell[c]{Kalman Filter \\ (Motion prediction)} & 0 & $\approx 0$ & 0.5 \\
        \midrule
        \makecell[c]{Set Transformer \\ (Scanpath selection)} & 0.56M & 0.01 & 3.0 \\
        \midrule
        \textbf{Total} & \textbf{5.66M} & \textbf{6.63} & \textbf{64.7} \\
        \bottomrule
    \end{tabular}
\end{table}

\section{Experiment Details and Results}
\subsection{Downstream Models (Main Paper Sec.~4.2)}
For object tracking, we use the MixFormerV2-Base model~\cite{cui2023mixformerv2}. We set the search area scale to 5, the update interval to 10, and the online size to 1. We keep all hyperparameter settings consistent across all object tracking evaluations.

For scene text recognition, we use the DeepSolo model~\cite{ye2023deepsolo}, finetuned on the ICDAR-15 Dataset~\cite{zhou2015icdar}. We intentionally disable automatic resizing and upsampling to a fixed resolution before the text detection and recognition pipeline for all experiments, as we compare the performance of our foveated imaging pipeline under both limited acquisition and processing bandwidth.

For robotic manipulation, we use the ALOHA Action Chunking Transformer (ACT)~\cite{zhao2023learning}. We set the ACT's internal policy seed to 0 and report results across 100 random environment configurations in which the objects vary in position.

\paragraph{Combining FFoV and ROI streams.} For object tracking and scene text recognition, the Full Field-of-View (FFoV) and ROI streams are fed \emph{separately} into the downstream perception models. For robotic manipulation, the downstream ACT model is trained to consume a single image per view, so we upsample the FFoV frame and overlay the high-resolution ROI at the corresponding location before feeding the composite frame to the model. We hypothesize that providing sharp, task-relevant content at the ROI supplies consistent visual attention cues that allow the frozen downstream model to maintain performance without retraining. Fine-tuning downstream perception models on mixed-resolution inputs remains future work.

\subsection{Alternative Acquisition Baselines (Main Paper Sec.~4.3)}
We compare against frame-skipping as a representative temporal downsampling baseline. An alternative temporal downsampling strategy that increases per-frame exposure yields substantially lower performance than frame-skipping ($0.098$, $0.071$, $0.00\mid 0.18$ for the three Main Paper Table~1 tasks under the same bandwidth) due to motion blur.

Prior foveation methods discussed in Main Paper Sec.~2.1 all operate \emph{post}-acquisition on already-captured full-resolution images or video. For example, Killick et al.~\cite{killick2023foveation} predict scanpaths on pre-captured images, and video-based methods such as AdaFocus~\cite{wang2024uni} rely on global video features that are unavailable at acquisition time. Adapting such methods for predictive, acquisition-time foveated imaging would require significant architectural changes. Main Paper Table~1 compares against foveation baselines that are directly applicable at acquisition time, and the ablation studies (Main Paper Sec.~4.5) further analyze the design choices within our policy.

\subsection{Additional Simulation Results (Main Paper Sec.~4.4)}
\begin{figure*}[!t]
    \centering
    \includegraphics[width=.76\linewidth]{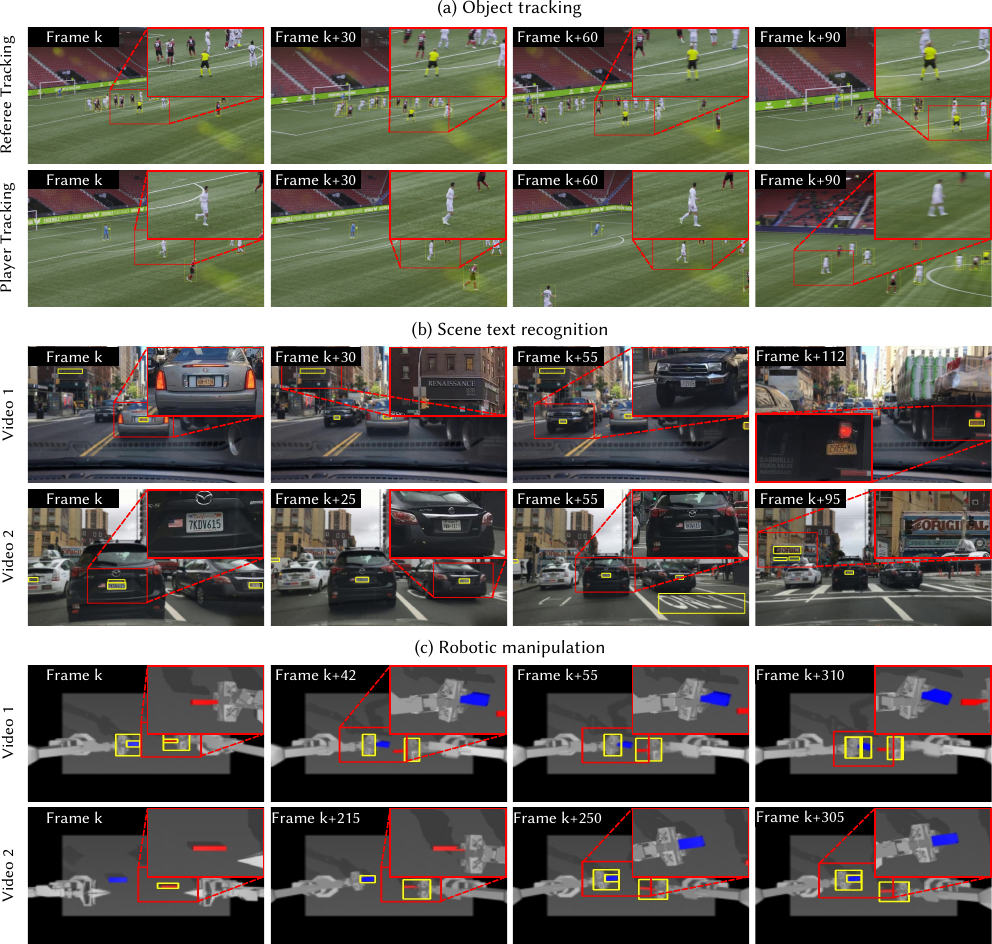}
    \caption{\textbf{Additional qualitative results for simulated video tasks.} We show additional examples of our foveated imaging framework across diverse scenes for object tracking, scene text recognition, and robotic manipulation, demonstrating consistent performance improvements over task-agnostic baselines.}
    \label{fig:supp_sim_results}
\end{figure*}

For the object tracking task, the visual conditioning information consists of global visual features from the initial object location and the ground-truth (GT) bounding box. Varying this conditioning enables our scanpath selection policy to select different objects of interest. Table~\ref{tab:additional_soccer} summarizes quantitative results of our approach for different subjects in soccer tracking. We include qualitative visualizations of smooth-pursuit scanpaths for various tracked subjects in our supplemental video. Our method consistently outperforms relevant baselines at comparable bandwidth. Fig.~\ref{fig:supp_sim_results} shows additional qualitative results across more diverse scenes for all three tasks. Across these scenes, our policy produces smooth-pursuit scanpaths for object tracking, rapid saccading scanpaths for scene text recognition, and a mixture of both for robotic manipulation.

\begin{table*}[!t]
    \caption{\textbf{Additional results on policy-based foveated perception for object tracking.} We compare the downstream object tracking task performance for different classes of objects, including the soccer ball, the referees, and the players. Our approach performs best among relevant baselines at comparable bandwidth across all tracking scenarios.}
    \label{tab:additional_soccer}
    \centering
    \begin{tabular}{c c c | c | c c c}
        \toprule
        Tracking subject & Metric & \makecell[c]{Full-resolution \\ ($8\times$ bandwidth)} & GT Oracle & \makecell[c]{Spatial \\ downsampling} & \makecell[c]{Temporal \\ downsampling} & \textbf{Foveated}\\
        \midrule
        Soccer ball & IoU $\uparrow$ & 0.281 & 0.405 & 0.122 & 0.148 & \textbf{0.283}\\
        \midrule
        Main referee & IoU $\uparrow$ & 0.494 & 0.602 & 0.474 & 0.456 & \textbf{0.583}\\
        \midrule
        Random player & IoU $\uparrow$ & 0.457 & 0.552 & 0.268 & 0.402 & \textbf{0.513}\\
        \bottomrule
    \end{tabular}
\end{table*}

\paragraph{Performance under different pixel budgets.}
To test how performance varies with bandwidth budget, we repeat the Main Paper Table~1 comparisons at higher and lower pixel budgets. Results are summarized in Table~\ref{tab:budget_sweep}. Our approach maintains the best performance across all budgets for all tasks, with the largest advantage occurring when conventional downsampling discards the most critical spatiotemporal details.

\begin{table*}[!t]
    \caption{\textbf{Pixel-budget sensitivity.} Our method outperforms spatial and temporal downsampling baselines at higher and lower pixel budgets across all three tasks. Rows marked with $^{\dagger}$ correspond to the pixel budgets reported in Main Paper Table~1 and are reproduced here for direct comparison against the additional budgets evaluated in this supplement.}
    \label{tab:budget_sweep}
    \centering
    \begin{tabular}{l c c c c}
        \toprule
        Task & \makecell[c]{Pixel \\ budget} & \makecell[c]{Spatial \\ downsampling} & \makecell[c]{Temporal \\ downsampling} & \textbf{Ours}\\
        \midrule
        Object Tracking & $1/4$ & 0.212 & 0.159 & \textbf{0.287}\\
        Object Tracking & $1/8^{\dagger}$ & 0.122 & 0.148 & \textbf{0.283}\\
        Object Tracking & $1/16$ & 0.048 & 0.144 & \textbf{0.276}\\
        \midrule
        Text Recognition & $1/4$ & 0.146 & 0.283 & \textbf{0.290}\\
        Text Recognition & $1/8^{\dagger}$ & 0.067 & 0.248 & \textbf{0.264}\\
        Text Recognition & $1/16$ & 0.023 & 0.173 & \textbf{0.227}\\
        \midrule
        Robotic Manipulation & $1/8$ & 0.12 $\mid$ 0.54 & 0.08 $\mid$ 0.38 & \textbf{0.13 $\mid$ 0.59}\\
        Robotic Manipulation & $1/16^{\dagger}$ & 0.10 $\mid$ 0.51 & 0.07 $\mid$ 0.30 & \textbf{0.12 $\mid$ 0.57}\\
        Robotic Manipulation & $1/32$ & 0.04 $\mid$ 0.38 & 0.02 $\mid$ 0.20 & \textbf{0.08 $\mid$ 0.40}\\
        \bottomrule
    \end{tabular}
\end{table*}

\section{Hardware Prototype}
\subsection{200~MP Foveated Imaging Prototype (Main Paper Sec.~5)}
Ultra-high-resolution sensors such as the 200~MP Samsung ISOCELL HP2~\cite{SamsungISOCELLHP2} are physically bandwidth-limited: the HP2 is capped at 15~fps at full resolution, with higher frame rates achievable only through pixel binning that reduces spatial resolution to 50~MP or 12.5~MP.

In our experimental setup, the prototype's hardware further restricts FFoV readout to 7.5~fps while the sensor captures at 30~fps. In addition, while the sensor is saving dual-stream raw Bayer frames at 30~fps, the sensor-captured frames are not accessible via the sensor's Python API. Therefore, we re-train both scanpath selection policies (for object tracking and scene text recognition) with $T_o = 10$ and $T_p = 200$ frames. Other hyperparameters are kept consistent with each policy's simulated-experiment training.

During real-world sensor captures, we capture up to 105 raw Bayer frames in dual-stream mode (both the FFoV and ROI streams have a 30~fps frame rate) while the sensor executes the ROI predictions from our foveated imaging pipeline. We use the replanning interval $T_a = 180$ in our real experiments. Without the sensor readout latency, our pipeline operates with a frame delay of about $12.8$ frames in the setting of $T_o = 4$ frames and $T_p = 16$ frames used in our simulated experiments. In real-world captures, our policy operates with a frame delay of about $31.3$ frames in the setting of $T_o = 10$ frames and $T_p = 200$ frames. Adding the sensor readout latency, the entire foveated imaging pipeline operates with approximately $60$ frames delay in this setting, well below the receding control horizon of $T_a = 180$ frames.

\subsection{Comparison to Samsung ISOCELL Zoom Anyplace}
Samsung ISOCELL Zoom Anyplace~\cite{SamsungISOCELLZoomAnyplace} is a user-driven, single-object tracking feature supported on the same ISOCELL HP2 sensor we use for our prototype. In contrast, our framework is neither user-driven nor hard-coded to a single task: it autonomously learns sensor attention from downstream perception objectives and generalizes across object tracking, scene text recognition, and robotic manipulation. Our contribution is not an open-source reimplementation of proprietary firmware on specific hardware, but rather an intelligent sensing framework that is sensor-agnostic, task-general, and extensible to other high-bandwidth sensing modalities, multi-camera systems, or closed-loop robotic pipelines.

\end{document}